\pdfoutput=1

\documentclass[11pt]{article}

\usepackage{ACL2023}
\usepackage{graphicx}
\usepackage{booktabs} 
\usepackage{threeparttable} 
\usepackage{multirow}
\usepackage{amssymb} 
\usepackage{bm}
\usepackage{array}
\usepackage{times}
\usepackage{makecell}
\usepackage{latexsym}
\usepackage{graphicx} 
\usepackage{algorithm}
\usepackage{algpseudocode}
\usepackage{amsmath} 
\usepackage[utf8]{inputenc}
\usepackage{graphicx}
\usepackage{booktabs}
\usepackage{tabularx}
\usepackage{amssymb}
\usepackage{xcolor}
\usepackage{xspace}
\usepackage[T1]{fontenc}

\usepackage[utf8]{inputenc}

\usepackage{microtype}

\usepackage{inconsolata}

\newcommand{\eg}{\emph{e.g.},\xspace}
\newcommand{\ie}{\emph{i.e.},\xspace}
\newcommand{\etc}{etc.\xspace}

\newcommand\figref[1]{Fig.~\ref{#1}}

\newcommand\tabref[1]{Table~\ref{#1}}

\newcommand\secref[1]{Sec.~\ref{#1}}

\newcommand{\fakeparagraph}[1]{\vspace{1mm}\noindent\textbf{#1.}}

\newcommand{\sysname}{\textsc{Pecola}\xspace }

\newcommand{\modelname}{\textsc{Pecola}\xspace}

\title{Does \textsc{DetectGPT} Fully Utilize Perturbation? Bridging Selective Perturbation to Fine-tuned Contrastive Learning Detector would be Better}

\author{Shengchao Liu\textsuperscript{$1$}, Xiaoming Liu\textsuperscript{$1,\ast$}, Yichen Wang\textsuperscript{$1$},  Zehua Cheng\textsuperscript{$1$}, Chengzhengxu Li\textsuperscript{$1$}, \\\textbf{Zhaohan Zhang\textsuperscript{$2$},  Yu Lan\textsuperscript{$1$},  Chao Shen\textsuperscript{$1$}} \\
        \textsuperscript{1}Faculty of Electronic and Information Engineering, Xi'an Jiaotong University\\ 
        \textsuperscript{2}Queen Mary University of London \\
        \texttt{
        \{liusc, yichen.wang, czh2022, czx.li\}@stu.xjtu.edu.cn
        }\\
        \texttt{
        \{xm.liu, ylan2020, chaoshen\}@xjtu.edu.cn, zhaohan.zhang@qmul.ac.uk
        }\\
        }

\begin{document}
\maketitle
\renewcommand{\thefootnote}{\fnsymbol{footnote}}
\footnotetext[1]{Corresponding author}
\renewcommand{\thefootnote}{\arabic{footnote}}
\begin{abstract}
The burgeoning generative capabilities of large language models (LLMs) have raised growing concerns about abuse, demanding automatic machine-generated text detectors.
DetectGPT \cite{mitchell2023detectgpt}, a zero-shot metric-based detector, first introduces perturbation and shows great performance improvement. 
However, in DetectGPT, the random perturbation strategy could introduce noise, and logit regression depends on the threshold, harming the generalizability and applicability of individual or small-batch inputs. 
Hence, we propose a novel fine-tuned detector, \modelname{}, bridging metric-based and fine-tuned methods by contrastive learning on selective perturbation.
Selective strategy retains important tokens during perturbation
and weights for multi-pair contrastive learning.
The experiments show that \modelname{} outperforms the state-of-the-art (SOTA) by 1.20\% in accuracy on average on four public datasets.
And we further analyze the effectiveness, robustness, and generalization of the method. \footnote{The code and datasets are released at \url{https://github.com/lsc-1/Pecola}.}
\end{abstract}

\section{Introduction}

Machine-generated text (MGT) detection is to discriminate MGT from human-written texts (HWT), preventing abuse of large language models (LLMs), including academic misconduct \cite{vasilatos2023howkgpt}, spam synthesis \cite{dou2020robust}, untrustworthy news \cite{zellers2019defending}, \etc{}
Currently, existing MGT detection methods can be mainly classified into two categories \cite{wu2023survey, wang2024stumbling}, \ie fine-tuned methods \cite{liu-etal-2023-coco,hu2023radar,verma2023ghostbuster, OpenAI2023MGTCls,mao2024raidar} and zero-shot metric-based methods \cite{gehrmann2019gltr, mitchell2023detectgpt, yang2023dna, bao2024fastdetectgpt,wu2023llmdet}. 
In general terms, fine-tuned detector methods can achieve better accuracy than zero-shot metric-based methods, especially generalizable to black-box generators, but are more costly during data collection, fine-tuning, and running, in most cases. On the other hand, zero-shot metric-based methods show better interpretability than fine-tuned ones.

\begin{figure}[t]
  \centering
  \includegraphics[width=\columnwidth]{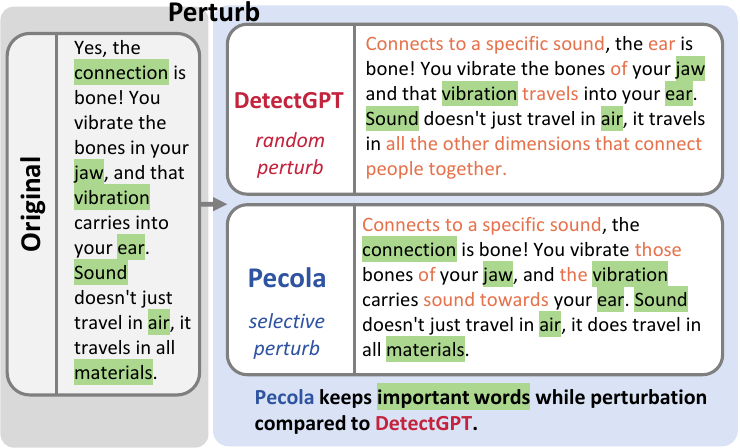}
  \caption{Example of the selective strategy perturbation of \modelname{}, which prevent modifying important tokens (in green). Orange tokens are the perturbed texts.}
  \label{fig:perturb} 
  \vspace{-15pt}
\end{figure}


DetectGPT \cite{mitchell2023detectgpt}, as an unsupervised zero-shot metric-based method, first introduces perturbation in MGT detection. 
Specifically, it applies random masking to the original input sample and uses T5 \cite{raffel2020exploring} to fill in.
It posits that minor perturbations of MGT tend to have lower log probability under the base model than the original sample.
The introduction of perturbation in DetectGPT surpasses the vanilla log-probability-based method \cite{gehrmann2019gltr} in white-box settings. 

However, DetectGPT still has three significant defects: 
(\textit{i}) DetectGPT's reliance on the logit regression module's threshold compromises its generalization in zero-shot settings and limited to large batch input, failing on individual inputs.
(\textit{ii}) DetectGPT does not fully utilize the perturbation. As a metrics-based method, it only considers the probability difference caused by perturbation, which is overly simplified and slightly indistinguishable. Perturbation should indeed be a stronger augment that carries implicit language pattern information.
(\textit{iii}) DetectGPT perturbs the original sample randomly and unrestricted, which could introduce more noise and negatively impact the performance \cite{kim2022alp}. For example, \citet{liu-etal-2023-coco} find entity-relationship plays a role in the detection, which might be destroyed in random perturbation of DetectGPT.

In this paper, we thus propose a \textbf{Pe}rturbation-based \textbf{Co}ntrastive \textbf{L}e\textbf{a}rning model, \sysname, for MGT detection, toward the defects via two stages, \ie Selective Strategy Perturbation (\secref{sec:Per}) and Token-Level Weighted Multi-Pairwise Contrastive Learning (\secref{sec:mcl}).
\textbf{Firstly}, Selective Strategy Perturbation is a token-level rewriting method with restrictions on modifying important texts \cite{campos2020yake} to reduce noise. 
The motivation is to simulate the human behavior of modification \cite{verma2017extractive,fetaya2019understanding,wang2019generalized}. 
The perturbation strategy consists of token removal and substitution, as shown in \figref{fig:perturb}.
The experiments show that the Selective Strategy Perturbation method can improve the performance of both metrics-based (\ie{} DetectGPT) and model-based methods.
\textbf{Secondly}, we propose a Multi-Pairwise Contrastive Learning model to process the perturbed texts.
Different from the logit regression module in DetectGPT, the trained model is generalizable without any threshold setting, and it can deal with individual inputs.
Moreover, by utilizing multi-pairwise contrastive learning, the model could better utilize perturbation to focus on the language pattern gap between HWT and MGT. 
The importance weight from the perturbation stage is also reused as contrastive learning weight.
Notably, by using contrastive learning, \modelname{} is a strong few-shot fine-tuning method, which effectively bridges and integrates metric-based and fine-tuned detector categories. 
\textbf{Finally}, extensive experiments show \modelname{} is significantly superior to baseline and SOTA methods on four datasets, 
\modelname{} improves by 1.20\% to SOTA on average under few-shot settings, surpassing the latest methods by 3.84\% among metric-based detectors and by 1.62\% among fine-tuned detectors.
Further experiments show that \modelname{} is also better at generalization, robustness, and effectiveness.

Our contributions are summarized as follows:

\begin{itemize}
    \item \textbf {Selective Perturbation}: Based on our analysis of various selective perturbation strategies,  we propose a novel method considering token importance, which reduces the noise and benefits to both supervised and unsupervised approaches. 
    \item \textbf{Bridge Metric and Model-based Detectors}: 
    We utilize a novel fine-tuned contrastive learning module to replace the logit regression of DetectGPT (metric-based), which frees the detector from setting the threshold, enables it to deal with individual input, and can be generalizable and effective on the few-shot setting by contrasting perturbed texts with origin ones.

    \item \textbf{Outperformance}: Our detector \modelname{} outperforms all eight compared models on four public datasets. And \modelname{} is more robust to the choice of base model and filling model. Furthermore,  we prove its generalization ability across domains and generators of data.
\end{itemize}

\section{Related Work}
\fakeparagraph{Machine-generated Text Detection} 
While fine-tuned detectors have proven effective for MGT detection \cite{wahle2022large, hu2023radar}, the requirement for annotated datasets poses a significant challenge due to the proliferation of unchecked, high-quality generated texts. 
To address this challenge, DetectGPT \cite{mitchell2023detectgpt} and Fast-DetectGPT \cite{bao2024fastdetectgpt} have demonstrated strong performance in white-box zero-shot settings.
Similarly, CoCo \cite{liu-etal-2023-coco} is designed to detect MGT with low resource annotations, utilizing a coherence-based contrastive learning model. 
Moreover, SeqXGPT \cite{wang2023seqxgpt} utilize log probability lists from white-box LLMs as features; Sniffer \cite{DBLP:journals/corr/abs-2402-09199} and GPT-Who \cite{venkatraman2023gpt} place more emphasis on tracing the origin of MGT.
Recently, watermarking \cite{kirchenbauer2023watermark} is introduced to mitigate the risk associated with unchecked MGTs by embedding imperceptible signals within text outputs during generation.
In contrast to previous methods, our approach integrates data perturbation with contrastive learning, placing particular emphasis on reducing reliance on mask-filling models and enhancing performance in few-shot scenarios.

\begin{figure*}
    \centering
    \includegraphics[width=\textwidth]{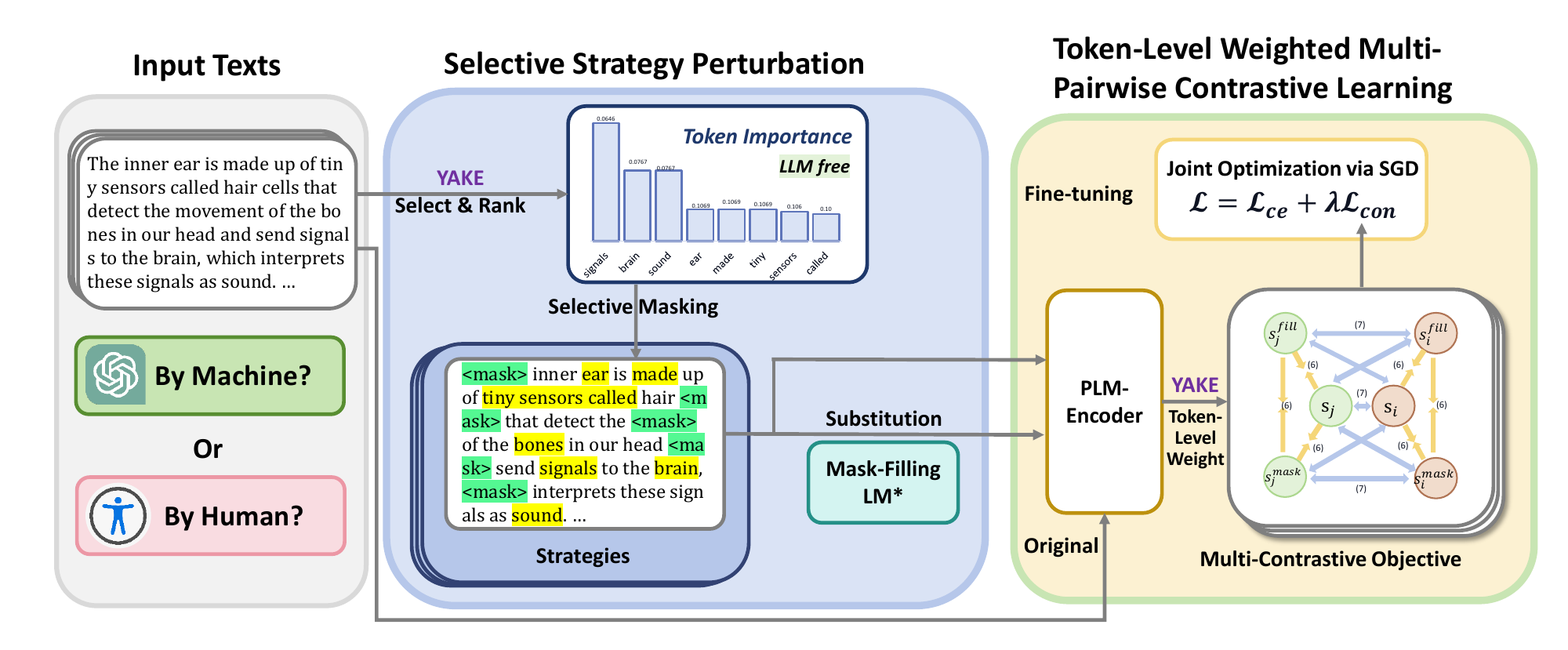}
    \caption{Overview of \modelname{}.
    In the Selective Strategy Perturbation stage (\secref{sec:Per}), we use the YAKE algorithm to score token importance and then selective masking based on probability. Then, we fill in the masks with a mark-filling language model.  In the Contrastive Learning stage (\secref{sec:mcl}), we design a multi-pairwise method with token-level weights also from tokens importance. Yellow arrows represent attraction and blue ones represent repulsion. The model is optimized by combining cross-entropy (CE) loss \(\mathcal{L}_{\text{ce}}\)  and contrastive loss \(\mathcal{L}_{\text{con}}\).
    * Our method, different from DetectGPT, is generalizable on any mask-filling language model. 
    }
    \label{fig:duco}
\end{figure*}

\fakeparagraph{Perturbation} 
Data perturbation methods find frequent application in text classification tasks \cite{gao2023mask,shum2023automatic}, which is commonly employed through the technique of consistency regularization \cite{xie2020unsupervised,chen2020mixtext}.
Nevertheless, in MGT detection, previous perturbation methods have exhibited certain limitations. 
For instance, they often resort to randomly selecting target tokens for synonym replacement \cite{wang2018switchout}, deletion, insertion \cite{wei2019eda}, rewriting by LLMs \cite{mao2024raidar}, 
and fine-tuning pre-trained language models (PLMs) to fill text spans of variable lengths \cite{gao2023mask}. 
While these methods do enhance text diversity, the indiscriminate replacement of tokens without guided rules can lead to the generation of less reliable texts.
\citet{wang2024stumbling} utilize perturbations as stress test approaches for the robustness of MGT detectors to show their loopholes.
These limitations motivate us to devise data perturbation methods tailored for MGT detection. 
Our approach, with selective perturbation, aims to better represent meaningful recombination spaces while preserving the inherent semantic features of the text, ultimately enhancing the diversity of samples.

\fakeparagraph{Constrastive Learning}
Contrastive learning is an effective solution method to the issues that solely relying on cross-entropy classification loss would lead to a lack of robustness and suboptimal generalization \cite{tack2020csi,hu2023radar}. 
In limited labeled data task \cite{gunel2020supervised}, introduce a robust contrastive learning method to capture the similarities between the same instances in the representation space while separating those of different classes.
Similarly, out-of-distribution (OOD) usually leads to severe semantic shift issues during inference, prompting another approach based on margin contrastive learning \cite{zhou2021contrastive}. 
Differently, our method focuses more on the changes of the rephrase space in data distribution after perturbation, and strives to reduce reliance on the mask-filling models in few-shot learning.

\section{Methodology}
As shown in \figref{fig:duco}, the workflow of \modelname{} mainly consists of two stages: Selective Strategy Perturbation and Supervised Contrastive Learning, which joined the advantage of metric-based and model-based detection methods, respectively.

\subsection{Selective Strategy Perturbation\label{sec:Per}}

In this work, we present a token-level selective strategy perturbation method to relieve the information loss caused by the random masking used in DetectGPT.
Our approach involves adapting the mask-selection probability for each text token based on its importance, thus generating perturbed inputs with strategically placed masks.
Additionally, we harness LLMs to populate the masks, creating filled perturbation inputs. This step effectively introduces a diverse range of perturbation information into our detection model.

\fakeparagraph{Token Importance Assessment}
To accurately assess the significance of tokens within the text and mitigate information loss stemming from random masking, we expand upon the YAKE algorithm \cite{campos2020yake} to operate at the token level.
The YAKE algorithm builds upon certain assumptions \cite{machado2009universal}, which posit that the importance of a candidate word decreases as the richness of the vocabulary surrounding it increases. 
This fundamental assumption remains applicable when processing text at the token level, \ie token importance assessment.

Specifically, considering a training set $S$ comprising $i$ inputs, for each text input $s_i \in S$ containing $n$ tokens (\ie $s_i = \{ e_i^{1}, e_i^{2}, \ldots, e_i^{n} \}$), we employ the YAKE algorithm to compute a score for each token $e$. Tokens with scores falling below the specified threshold $\alpha$ are then incorporated into the set of important tokens $K_i$:
\begin{align} \label{eq:1}
    K_i = \begin{cases} 
  K_i \cup \left \{e_i^{n}\right \}, & \text{if } Score(e_i^{n}) < \alpha \\
  K_i, & \text{otherwise}
\end{cases}
,
\end{align}
where $Score(e_i^{n})$ represents the YAKE score calculated by token $e_i^{n}$. The higher the score, the lower the importance of the token $e_i^{n}$ in $s_i$.

\noindent\textbf{Mask Position Selection}. 
After getting the important tokens set $K_i$ of each text input $s_i$, we use special token \texttt{[MASK]} to replace some of the tokens in the text input to construct masked perturbation input $s_i^{mask}$. In order to relieve the information loss caused by masking perturbation, we add regularization to the vanilla random masking method and use a selective masking strategy to prevent important tokens from being masked.

Given an input text $s_i = \left \{ e_i^{1} , e_i^{2},\dots, e_i^{n}\right \} $, we use the selective masking strategy to traverse each token and determine whether to mask it based on the token's importance. The probability of token $e_i^{n}$ being masked is specifically defined as: 
\begin{align}
    P_i^{n} = \mathbf{1} _{[e_i^{n} \notin K_i]}P
,
\end{align}
where $P$ is the mask ratio, and $\mathbf{1} _{[e_i^{n} \notin K_i]}$ represents an indicator function with a value of $1$ if and only if the condition $e_i^{n} \notin K_i$ is satisfied, otherwise, it is $0$. Then we gather all masked perturbation inputs $\{s_1^{mask}, ..., s_i^{mask}\}$ and include them in the training set to give the model masked perturbation, improving model robustness.

\noindent\textbf{Mask-Filling}.
Additionally, we utilize PLMs, \eg T5 \cite{raffel2020exploring} or RoBERTa \cite{liu2019roberta} \etc, to fill the masked perturbation inputs and create the filled perturbation inputs $\{s_1^{fill}, \dots, s_i^{fill}\}$. Similar to the above, we include all filled perturbation inputs in the training set and obtain the final training set $S =  \{ s_1, \dots, s_i, s_1^{mask}, \dots, s_i^{mask}, s_1^{fill} \dots, s_i^{fill}  \} $.

\subsection{Token-Level Weighted Multi-Pairwise Contrastive Learning\label{sec:mcl}}
\noindent\textbf{Importance-based Feature Reconstruction}.
Existing MGT methods \cite{liu-etal-2023-coco} often uniformly extract all token information in the text, ignoring the huge impact of a few important tokens on the detection model. In this work, we reconstruct the token feature extracted by PLM according to the importance of the token in the input text, allowing the detection model to focus more on important token information.
We assign adaptive weights to all tokens in the input:
\begin{equation}
\label{eq:eq5}
w_i^{n}=\left\{\begin{matrix}
 1-Score(e_i^{n}), &\text{if } e_i^{n}\in K_i\\
0,&\text{otherwise}
\end{matrix}\right.
,
\end{equation}
where $w_i^{n}$ represents the assign adaptive weight of the $n$-th token of the $i$-th input in the training set. After that, we use the last hidden layer embedding of the outputs in the base PLMs to extract input features:
\begin{equation}
H_i=\mathrm{PLM}(s_i)
,
\end{equation}
where $H_i$ contains the features of all tokens in the input $s_i$, \ie $H_i = \{h_i^{1}, h_i^{2}, \dots, h_i^{n}\}$. We use the weight of the corresponding token to reconstruct its features:
\begin{equation}
h_i^{n}=h_i^{n}(1+w_i^{n})
.
\end{equation}
By using feature reconstruction, we assign more weight to important tokens. This allows our detection model to concentrate on the characteristic information of these important tokens.

\noindent\textbf{Multi-Pairwise Contrastive Learning}.
Considering that existing works \cite{gunel2020supervised,zhou2021contrastive,liu-etal-2023-coco} mainly concentrate on single-input feature learning while overlooking input correlations, we introduce contrastive learning into MGT. It enables \modelname{} to discern the distinct featurinputes of variously labeled data, more accurately capture input features, and significantly enhance performance in few-shot setting.

Given a batch training data $\{s_i\}_{i=1}^{M}$, where $M$ is the batch size, we calculate the positive class contrastive loss and negative class contrastive loss on the last hidden layer embedding of the first token output $h_i^{1}$ from the base PLM: 
\begin{equation}
\mathcal{L}_{\text{pos}} = \sum_{i=1}^{M}\frac{1}{|P_t(i)|}\sum_{p \in P_t(i)} \|(h_i^{1} - h_{p}^{1}) \|^2 
\vspace{-15pt}
,
\end{equation}

\begin{equation}
\footnotesize
\mathcal{L}_{\text{neg}} = \sum_{i=1}^{M}  \frac{1}{|N_t(i)|} \sum_{n \in N_t(i)} \max \left( 0, \xi - \|(h_i^{1} - h_{n}^{1}) \|^2 \right)
,
\end{equation}
where $P_t(i)$ represents the samples with the same label as the $i$-th sample in the batch, and $N_t(i)$ represents the ones with different labels as the $i$-th sample. 
And $\xi$ is the maximum $L_{2}$ distance between pairs of inputs from the same class in the batch of training data:
\begin{equation}
\xi = \max_{i=1}^{M} \max_{\substack{p \in P_t(i)}} \| h_i^{1} - h_{p}^{1} \|^2
.
\end{equation}
This adaptive margin ensures that the model is steered to maintain discriminative embeddings despite data perturbation during training. Then we get the following contrastive loss as:
\begin{equation}
\mathcal{L}_{\text{con}} = \frac{1}{M} (\mathcal{L}_{\text{pos}} + \mathcal{L}_{\text{neg}})
.
\end{equation}
For supervised learning tasks, we utilize the cross-entropy classification loss $\mathcal{L}_{ce}$ to train our detection model. By adjusting the weight $\lambda$ to balance the impact of various losses on the model, our total loss is given by the following:
\begin{equation}
\mathcal{L}= \mathcal{L}_{\text{ce}} + \lambda \mathcal{L}_{\text{con}}
.
\end{equation}

\begin{table*}[ht]
\centering
\renewcommand\arraystretch{1.22}
\resizebox{\textwidth}{!}{
\begin{tabular}{cccccccccccc}
\toprule
{\textbf{Dataset}} 
& {\textbf{Metric}}
& {\textbf{Shot}}
& {\textbf{\textit{RoBERTa}}} 
& {\textbf{\textit{GLTR}$^\dag$}} 
& {\textbf{\textit{CE+SCL}}} 
& {\textbf{\textit{CE+Margin}}} 
& {\textbf{\textit{IT:Clust}}}
& {\textbf{\textit{CoCo}}$_{\textbf{*}}$}
& {\textbf{\textit{DetectGPT}$_{\textbf{*}}^\dag$}} 
& {\textbf{\textit{Fast-Detect.$_{\textbf{*}}^\dag$}}} 
& {\textbf{\textit{\modelname{}}}}  \\
\toprule
            \multirow{8}{*}{\textbf{\rotatebox{90}{Grover}}}
            & \multirow{4}{*}{\textit{Acc}}&32 
            & 48.83$_{\text{10.31}}$ & 56.61 & 55.86$_{\text{4.43}}$ & 56.79$_{\text{3.31}}$ & 41.57$_{\text{3.58}}$ &51.60$_{\text{8.42}}$ & 55.02 & 56.06
            & \textbf{59.03}$_{\text{\textbf{1.63}}}$ \\
            
            &&64 & 56.88$_{\text{3.03}}$ & 56.61 & 57.57$_{\text{2.63}}$ & 58.92$_{\text{2.17}}$ & 46.45$_{\text{2.20}}$ & 58.27$_{\text{10.21}}$ & 54.61 &  60.33
            & \textbf{60.94}$_{\text{\textbf{1.56}}}$ \\
            
            &&128 & 59.28$_{\text{1.91}}$ & 58.48 & 60.33$_{\text{3.41}}$ & 60.44$_{\text{3.85}}$ & 50.72$_{\text{3.70}}$ & 58.97$_{\text{5.53}}$ & 55.78 & 60.33
            & \textbf{63.60}$_{\text{\textbf{1.71}}}$ \\
            
            &&512 & 70.39$_{\text{1.21}}$ & 62.26 & 72.38$_{\text{1.73}}$ & 72.15$_{\text{1.16}}$ & 56.08$_{\text{0.87}}$ & 70.07$_{\text{5.54}}$ & 55.56 & 62.50
            & \textbf{73.12}$_{\text{\textbf{0.84}}}$ \\
            \cline{3-12} 
    
            &\multirow{4}{*}{\textit{F1}}
            &32 & 44.13$_{\text{8.82}}$ & 52.77 & 51.56$_{\text{3.03}}$ & 53.21$_{\text{2.24}}$ & 40.79$_{\text{3.66}}$ & 47.33$_{\text{2.63}}$ & 51.09 & 56.67
            & \textbf{53.95}$_{\text{\textbf{0.94}}}$ \\
            
            &&64 & 52.88$_{\text{1.52}}$ & 52.77 & 53.39$_{\text{1.16}}$ & 54.99$_{\text{1.75}}$ & 46.10$_{\text{1.25}}$ & 44.70$_{\text{3.53}}$ & 48.07 &  57.92
            & \textbf{55.48}$_{\text{\textbf{1.35}}}$ \\
            
            &&128 & 54.69$_{\text{1.18}}$ & 54.47 & 55.74$_{\text{2.21}}$ & 55.54$_{\text{2.40}}$ & 51.37$_{\text{4.80}}$ & 51.44$_{\text{2.13}}$ & 53.78 & 54.89
            & \textbf{58.98}$_{\text{\textbf{1.58}}}$ \\
            
            &&512 & 64.49$_{\text{3.17}}$ & 57.11 & 67.02$_{\text{2.12}}$ & 66.25$_{\text{1.65}}$ & 51.80$_{\text{0.49}}$ & 65.15$_{\text{3.76}}$ & 53.32 & 61.29
            & \textbf{68.24}$_{\text{\textbf{1.64}}}$ \\
            
        \midrule
            \multirow{8}{*}{\textbf{\rotatebox{90}{GPT-2}}}
            &\multirow{4}{*}{\textit{Acc}}
            &32 & 70.53$_{\text{4.10}}$ & \textbf{75.99} & 69.32$_{\text{5.19}}$ & 70.00$_{\text{2.33}}$ 
            & 51.02$_{\text{1.66}}$ & 71.69$_{\text{7.07}}$ & 68.59 & 71.88
            & 75.42$_{\text{1.80}}$ \\
    
            &&64 & 74.41$_{\text{2.47}}$ & 75.76 & 73.77$_{\text{3.54}}$ & 74.04$_{\text{1.42}}$ & 54.32$_{\text{2.73}}$ & 73.20$_{\text{1.42}}$ & 71.12 &  71.88
            & \textbf{78.92}$_{\text{\textbf{1.14}}}$ \\
    
            &&128 & 79.77$_{\text{2.04}}$ & 75.77 & 80.18$_{\text{1.25}}$ & 80.93$_{\text{1.26}}$ & 59.66$_{\text{2.83}}$ & 79.44$_{\text{4.80}}$ & 71.74 &  71.88
            & \textbf{82.58}$_{\text{\textbf{0.49}}}$ \\
            
            &&512 & 84.07$_{\text{1.46}}$ & 75.86 & 84.76$_{\text{1.19}}$ & 84.89$_{\text{1.17}}$ & 71.59$_{\text{3.23}}$ & 84.30$_{\text{0.58}}$ & 71.74 &74.06
            & \textbf{85.75}$_{\text{\textbf{0.69}}}$ \\
            \cline{3-12}
            
            &\multirow{4}{*}{\textit{F1}}
            &32 & 66.57$_{\text{5.09}}$ & 72.45 & 64.89$_{\text{8.13}}$ & 69.89$_{\text{2.38}}$ & 48.45$_{\text{3.72}}$ & 71.19$_{\text{11.05}}$ & 65.50 & 70.00
            & \textbf{75.10}$_{\text{\textbf{1.99}}}$ \\
            
            &&64 & 73.91$_{\text{2.69}}$ & 70.87 & 72.32$_{\text{4.31}}$ & 73.94$_{\text{1.40}}$ & 53.87$_{\text{3.00}}$ & 69.79$_{\text{2.03}}$ & 66.58 & 70.97
            & \textbf{78.88}$_{\text{\textbf{1.17}}}$ \\
            
            &&128 & 79.49$_{\text{2.26}}$ & 71.16 & 80.00$_{\text{1.35}}$ & 80.79$_{\text{1.34}}$ & 59.48$_{\text{2.79}}$ & 76.10$_{\text{7.37}}$ & 66.13 & 71.88
            & \textbf{82.54}$_{\text{\textbf{0.51}}}$ \\
            
            &&512 & 84.01$_{\text{1.52}}$ & 75.56 & 84.72$_{\text{1.25}}$ & 84.86$_{\text{1.24}}$ & 70.42$_{\text{4.26}}$ & 83.88$_{\text{0.79}}$ & 66.13 & 74.64
            & \textbf{85.72}$_{\text{\textbf{0.70}}}$ \\
        \midrule
            \multirow{8}{*}{\textbf{\rotatebox{90}{GPT-3.5}}}
            & \multirow{4}{*}{\textit{Acc}}
            &32 & 90.54$_{\text{{7.26}}}$ & 92.55 & 92.44$_{\text{3.19}}$ & 92.85$_{\text{2.44}}$ & 61.82$_{\text{4.30}}$ & 93.27$_{\text{1.44}}$ & 84.42 &89.10 
            & \textbf{95.80}$_{\text{\textbf{0.68}}}$ \\
            
            &&64 & 96.85$_{\text{0.84}}$ & 91.00 & 96.86$_{\text{1.67}}$ & 97.32$_{\text{0.58}}$ & 77.70$_{\text{6.92}}$ & 95.76$_{\text{1.52}}$ & 82.58 & 89.65
            & \textbf{98.01}$_{\text{\textbf{0.31}}}$ \\
            
            &&128 & 97.50$_{\text{1.24}}$ & 91.60 & 98.00$_{\text{0.46}}$ & 98.00$_{\text{0.18}}$ & 92.54$_{\text{4.01}}$ & 96.26$_{\text{0.89}}$ & 85.33 & 89.85
            & \textbf{98.06}$_{\text{\textbf{0.12}}}$ \\
            
            &&512 & 98.97$_{\text{0.18}}$ & 92.60 & 98.99$_{\text{0.80}}$ & 98.92$_{\text{0.28}}$ & 98.13$_{\text{1.20}}$ & 98.05$_{\text{0.47}}$ & 85.57 &90.62
            & \textbf{99.14}$_{\text{\textbf{0.15}}}$ \\
            \cline{3-12}
            
            &\multirow{4}{*}{\textit{F1}}
            &32 & 90.27$_{\text{7.77}}$ & 92.71 & 92.42$_{\text{3.20}}$ & 92.81$_{\text{2.49}}$ & 60.95$_{\text{4.67}}$ & 92.72$_{\text{1.54}}$ & 84.43 &  89.76
            & \textbf{95.80}$_{\text{\textbf{0.68}}}$ \\
            
            &&64 & 96.84$_{\text{0.84}}$ & 91.49 & 96.86$_{\text{1.67}}$ & 97.47$_{\text{0.30}}$ & 77.33$_{\text{7.31}}$ & 95.45$_{\text{1.54}}$ & 86.16 & 89.92
            & \textbf{98.01}$_{\text{\textbf{0.31}}}$ \\
            
            &&128 & 97.50$_{\text{1.24}}$ & 91.96 & 98.00$_{\text{0.46}}$ & 98.00$_{\text{0.18}}$ & 92.50$_{\text{4.07}}$ & 97.57$_{\text{0.92}}$ & 86.13 &  89.77
            & \textbf{98.06}$_{\text{\textbf{0.12}}}$ \\
            
            &&512 & 98.85$_{\text{0.40}}$ & 92.71 & 98.93$_{\text{0.21}}$ & 98.92$_{\text{0.28}}$ & 98.13$_{\text{1.20}}$ & 97.88$_{\text{0.50}}$ & 86.20 & 90.62
            & \textbf{99.14}$_{\text{\textbf{0.15}}}$ \\
            
        \midrule
            
            \multirow{8}{*}{\textbf{\rotatebox{90}{HC3}}}& \multirow{4}{*}{\textit{Acc}}
            &32 & 93.36$_{\text{1.50}}$ & \textbf{97.30} & 95.33$_{\text{1.81}}$ & 95.46$_{\text{1.71}}$ & 77.00$_{\text{8.05}}$ & 92.11$_{\text{1.71}}$ & 94.54 &  87.70
            & 97.19$_{\text{0.16}}$ \\
            
            &&64 & 96.97$_{\text{0.74}}$ & 98.13 & 97.81$_{\text{0.41}}$ & 97.81$_{\text{0.31}}$ & 91.69$_{\text{2.34}}$ & 95.50$_{\text{1.27}}$ & 95.03 &  88.87
            & \textbf{98.59}$_{\text{\textbf{0.14}}}$ \\
            
            &&128 & 97.56$_{\text{0.38}}$ & 98.29 & 98.17$_{\text{0.30}}$ & 98.14$_{\text{0.36}}$ & 95.43$_{\text{1.15}}$ & 97.57$_{\text{1.09}}$ & 95.10 &  88.87
            & \textbf{98.63}$_{\text{\textbf{0.32}}}$ \\
            
            &&512 & 98.85$_{\text{0.40}}$ & 98.31 & 98.93$_{\text{0.21}}$ & 98.99$_{\text{0.20}}$ & 97.98$_{\text{0.47}}$ & 98.58$_{\text{1.18}}$ & 95.13 & 90.62
            & \textbf{99.15}$_{\text{\textbf{0.11}}}$ \\
            
        \cline{3-12}
            
            &\multirow{4}{*}{\textit{F1}}
            &32 & 93.34$_{\text{1.52}}$ & \textbf{97.30} & 95.32$_{\text{1.82}}$ & 95.45$_{\text{1.72}}$ & 76.47$_{\text{8.77}}$ & 92.07$_{\text{1.56}}$ & 94.29 & 88.39
            & {97.19}$_{\text{0.16}}$ \\
            
            &&64 & 96.97$_{\text{0.74}}$ & 98.12 & 97.81$_{\text{0.41}}$ & 97.81$_{\text{0.32}}$ & 91.67$_{\text{2.34}}$ & 95.50$_{\text{1.19}}$ & 94.95 &  89.92
            & \textbf{98.59}$_{\text{\textbf{0.14}}}$ \\
            
            &&128 & 97.56$_{\text{0.38}}$ & 98.29 & 98.17$_{\text{0.30}}$ & 98.14$_{\text{0.36}}$ & 95.43$_{\text{1.15}}$ & 97.59$_{\text{1.05}}$ & 95.01 &  89.92
            & \textbf{98.63}$_{\text{\textbf{0.32}}}$ \\
            
            &&512 & 98.85$_{\text{0.40}}$ & 98.31 & 98.93$_{\text{0.21}}$ & 98.99$_{\text{0.20}}$ & 97.98$_{\text{0.47}}$ & 98.59$_{\text{1.16}}$ & 95.05 &  91.06
            & \textbf{99.15}$_{\text{\textbf{0.11}}}$ \\
            
\bottomrule
              
\end{tabular}
}

\caption{Comparison of \modelname{} to baseline methods in few-shot MGT detection. 
The results are average values of 10 runs with different random seeds. The subscript means the standard deviation (\eg $99.15_{0.11}$  means 99.15 ± 0.11). 
$\dag$ Zero-shot model-based methods' results are deterministic, so we do not report standard deviation. Also, these methods must have the white-box generator as the base model, which is different from the black-box settings of other model-based methods. Asterisk (\textbf{*}) denotes the latest SOTA method.
And we also conduct a more in-depth test on the entire training set in Appendix~\ref{app:fulldata}.
    }
\label{tab:ducomain}
\end{table*}

\section{Experiments} 

\subsection{Experiment Settings} 
To demonstrate the effectiveness of \modelname{}, we conduct extensive experiments on four open-source datasets under few-shot learning settings.

\fakeparagraph{Datasets}
\textbf{Grover} \cite{zellers2019defending}, generated by the transformer-based news generator Grover-Mega (1.5B);  
\textbf{GPT-2}, a webtext dataset provided by \citet{OpenAI2019GPT2Dataset} based on GPT-2 XL (1.5B);  
\textbf{GPT-3.5}, a news-style dataset constructed by CoCo \cite{liu-etal-2023-coco} using the text-DaVinci-003 model (175B);  
\textbf{HC3} \cite{guo2023close}, involving open domains, finance, healthcare, law, and psychology texts, composed of comparative responses from human experts and ChatGPT.

\fakeparagraph{Few-shot Learning Settings}
We randomly sample 32, 64, 128 and 512 samples from the original training set, while keeping the balance of machine and human categories. More details are provided in Appendix~\ref{app:details}.


\subsection{Comparison Models} 
We compare \modelname{} with both unsupervised and supervised MGT detection methods:

\noindent\textbf{\textit{RoBERTa}} \cite{liu2019roberta}, supervised methods via standard fine-tuning PLMs as classifiers. We use RoBERTa-base (125M).
\newcommand{\textcite}[1]{\citeauthor{#1}~\citeyear{#1}}

\noindent\textbf{\textit{GLTR}} \cite{gehrmann2019gltr}, 
a metric-based detector and based on next-token probability.
We follow the setting of \citet{guo2023close}, utilizing the Test-2 feature. For a fair comparison with fine-tuning methods, we first use the few-shot training samples to settle the threshold and adapt the fixed threshold in the test set.\footnote{The base model of \textit{GLTR} is chosen based on the generator of the dataset: for GPT-2 and Grover datasets, we use GPT-2 Small (124M); and for GPT-3.5 and HC3 datasets, we use GPT-J (6B) \cite{mesh-transformer-jax}, which is the best open-source model to simulate ChatGPT and GPT-3.5 empirically.}

\noindent\textbf{\textit{CE+SCL}} \cite{gunel2020supervised}, 
a fine-tuned detector, used in conjunction with the Cross-Entropy (CE) loss, exhibiting impressive performance in few-shot learning settings. 

\noindent\textbf{\textit{CE+Margin}} \cite{zhou2021contrastive}, a contrastive learning approach focuses on separating OOD instances from In-Distribution (ID) instances, aiming to minimize the L$_{\text{2}}$ distance between instances of the same label. We train the detector by combining CE loss.

\noindent\textbf{\textit{IT:Clust}} \cite{shnarch2022cluster}, a general text classification method that employs unsupervised clustering as an intermediate for fine-tuning PLMs, utilizing RoBERTa-base.

\noindent\textbf{\textit{CoCo}} \cite{liu-etal-2023-coco} utilizes coherence graph representation and contrastive learning to improve supervised fine-tuning methods in both inadequate and adequate data resource scenarios.

\noindent\textbf{\textit{DetectGPT}} \cite{mitchell2023detectgpt},
a zero-shot metric-based MGT detector,
using T5-large \cite{raffel2020exploring} to perturb texts.
Same as \textit{GLTR}, we fix the threshold.\footnote{For all four datasets (including HC3 and GPT-3.5 datasets), we use GPT-2 Small (124M) as the base model to calculate the likelihood. The reason is \citet{mireshghallah2023smaller} find that small model is better black-box detector for \textit{DetectGPT}.}

\noindent\textbf{\textit{Fast-DetectGPT}} \cite{bao2024fastdetectgpt}, an optimized zero-shot detector, building upon the foundation of DetectGPT, and utilizes a surrogate GPT-Neo (2.7B) \cite{black2022gpt} model for scoring.



\subsection{Performance Comparison}
As shown in Table~\ref{tab:ducomain}, \modelname{} surpasses the competitors on all datasets in the few-shot MGT detection task. 
Specifically, compared with the best competitor, \modelname{} achieves accuracy and F1-score improvement of 2.04\% and 1.42\%, 1.71\% and 2.55\% on Grover and GPT2 datasets. 
On GPT3.5 and HC3 datasets, \modelname{} still ensures 0.86\% and 0.68\%, 0.21\% and 0.22\% performance improvement with greater stability.
The results prove the effectiveness of \modelname{}, which integrates the advantage of unsupervised (perturbation for metric-based) and supervised (contrastive learning for model-based) MGT detection methods.

Moreover, the unsupervised learning methods tend to show better performance in extremely few shot scenarios. 
Unsurprisingly, unsupervised methods do not see a notable performance improvement with the increase in the number of training samples, which causes them to outperform on the fewest shot settings initially but soon be surpassed.  
As for the deception of generators, Grover appears to be the hardest to detect, while other models are relatively ``honest'' to detectors.
It might have originated from the adversarial training strategy of Grover, while the bulit-in detector module adversarially shifts the LLM's detectable features.
More interestingly, advanced language models show a weaker ability to cheat detectors.
Most detectors achieve around 98\% in accuracy on the GPT-3.5 and HC3 datasets, which is consistent with the conclusion from \citet{liu-etal-2023-coco, chen2023gpt}.
We hypothesize that the easy-to-detect nature may originate from the lack of semantics diversity in GPT-3.5 and ChatGPT as they use RLHF \cite{kirk2023understanding}. 

\subsection{Ablation Study}
To illustrate the effectiveness of the \modelname{} components, we do the ablation experiments on the Selective Strategy Perturbation stage and the Contrastive Learning stage on the 64-example GPT-2 dataset. We also demonstrate the Scalability of \modelname{} in Appendix~\ref{app:scalability}.
\begin{table}[ht]
\renewcommand{\arraystretch}{1.2}

\footnotesize %
\centering 
{
\begin{tabular}{llcccc}
\toprule
\multicolumn{1}{l}{\textbf{Method}} & \multicolumn{1}{c}{\textbf{\textit{Acc}}}&& \textbf{\textit{F1}} \\ \midrule
w/o. mask & \multicolumn{1}{c}{ 78.00$_{\text{1.40}}$}& &{ 77.93$_{\text{1.43}}$} \\ 
w/o. mask-fill &\multicolumn{1}{c}{ 77.78$_{\text{1.82}}$}& & { 77.72$_{\text{1.83}}$} \\ 
w/o. mask. CL$_{w}$ & \multicolumn{1}{c}{ 75.80$_{\text{2.22}}$}& &{ 75.23$_{\text{2.46}}$}\\ 
w/o. mask-fill. CL$_{w}$ & \multicolumn{1}{c}{ 75.56$_{\text{1.47}}$}& &{ 75.10$_{\text{1.73}}$}\\ 
w/o. CL$_{w}$ & \multicolumn{1}{c}{ 76.60$_{\text{1.69}}$} &&{ 76.22$_{\text{1.65}}$}\\ 
w/o. $w$ & \multicolumn{1}{c}{ 78.02$_{\text{1.56}}$} &&{ 77.93$_{\text{1.57}}$}\\ \midrule
\modelname{} & \multicolumn{1}{c}{ 78.92$_{\text{1.14}}$} &&{ 78.88$_{\text{1.17}}$}\\ \bottomrule
\end{tabular}
}
\caption{
Ablation study result of \sysname{}.
}
\label{tab:ablation}
\end{table}

\fakeparagraph{Ablation on Selective Strategy Perturbation} In \modelname{}, the data used for training primarily includes original texts, selected mask texts, and mask-filled texts. 
We remove each part of the data in training, \ie{} (\textit{i}) \textbf{w/o. mask}, refers to not using selected mask texts for training; (\textit{ii}) \textbf{w/o. mask-fill}, not using mask-filling texts for training. 

\fakeparagraph{Ablation on Contrastive Learning} It primarily investigates the impact of CE and contrastive loss.
(\textit{i}) \textbf{w/o. CL$_{w}$} refers to the model ablating weighted contrastive learning;
(\textit{ii}) \textbf{w/o. $w$} refers to the model including contrastive learning but ablating weight.

As demonstrated in \tabref{tab:ablation}, in scenarios employing only the CE loss, the Selective Strategy Perturbation method contributes to significant performance improvement.
Moreover, the introduction of weighting further enhances accuracy when compared to the direct use of margin loss.
It reveals the validation of bridging the metric-based and model-based detectors, \ie employing the Selective Strategy Perturbation method to evaluate the token importance for the multi-pairwise contrastive learning method.
Furthermore, within the overall framework, the removal of the select mask text results in a more rapid decrease in accuracy compared to the removal of the mask-filling text.
This finding substantiates that the Token-Level Weighted Multi-Pairwise Contrastive Learning method can better focus on the alterations in the rephrased space following the application of Selective Strategy Perturbation to the text. 

\subsection{Discussion and Analysis}

\subsubsection{Model Qualities}
We analyze the model qualities, including robustness and affinity in this section.
Here, we test on the 10,000-example GPT-2 test dataset, and the perturbation scale is set to 15\%.

\fakeparagraph{Analysis on Robustness} To validate the robustness of \modelname{}  in the few-shot learning settings, we apply four post hoc perturbation operations for each token in the test dataset randomly, \ie deletion, replacement, insertion, and repetition. 
As indicated in Table~\ref{tab:Perturbation}, for each perturbation method employed, our decline rate is consistently lower compared to the baseline RoBERTa. On average, \modelname{} maintains a 5.66\% higher accuracy and an 8.77\% superior F1-score.
Specifically, in the deletion method, where we introduce a 15\% random perturbation, it is noteworthy that the accuracy of \modelname{} decreases merely 1.64\%, underscoring its remarkable robustness.

\begin{table}[h]
\renewcommand{\arraystretch}{1.4}
\resizebox{\columnwidth}{!}{
\begin{tabular}{lccccccccc}
\toprule
\textbf{Model}& \multicolumn{2}{c}{\textbf{RoBERTa}}&&\multicolumn{2}{c}{\textbf{\modelname{}}} \\ \cline{2-3} \cline{5-6} 
\multicolumn{1}{l}{{\textbf{Metric}}} & \multicolumn{1}{c}{\textit{Acc}}& \textit{F1}&&\multicolumn{1}{c}{\textit{Acc}}&\textit{F1} \\  \midrule
Original& \multicolumn{1}{c}{74.41$_{2.47}$} & { 73.91$_{2.69}$}&&{ 78.92$_{1.14}$}&{ 78.88$_{1.17}$}\\  \midrule
Delete& {71.77$_{5.88}$}(-2.640)& {70.42$_{8.05}$}(-3.490)&& {77.28$_{1.79}$}(-1.640)& {77.06$_{2.03}$}(-1.820)\\ 
Repeat& {64.69$_{6.63}$}(-9.720)&{ 61.74$_{9.20}$}(-12.17)&& { 69.74$_{4.83}$}(-9.180)&  {67.87$_{6.24}$}(-11.01)\\ 
Insert& {50.75$_{0.67}$}(-23.66)&{ 36.44$_{1.60}$}(-37.47)&& { 57.61$_{2.52}$}(-21.31)& { 49.29$_{4.57}$}(-29.59) \\ 
Replace & {52.04$_{1.58}$}(-22.37)&{ 39.48$_{3.59}$}(-34.43)&& {57.25$_{2.21}$}(-21.67)& {48.89$_{3.93}$}(-29.99)\\ \midrule
Average& {59.81 (-14.60)} & 52.02 (-21.89)&& 65.47 (-13.45)& 60.78 (-18.10)\\ \bottomrule
\end{tabular}
}
\caption{
Model robustness to four perturbations.
}
\label{tab:Perturbation}
\end{table}

\noindent\textbf{Analysis on Affinity.}
Affinity pertains to alterations in data distribution resulting from perturbations, quantified by observing the fluctuations in accuracy. 
We demonstrate the superiority of the selective masking method over the random masking method using the Affinity metric, following the setting of DetectGPT. 
We applied a 15\% mask proportion with a span of 2 tokens on the test dataset and simultaneously employed T5-Large \cite{raffel2020exploring} as the mask-filling model.
We trained RoBERTa-base and \modelname{} on the 64-example GPT2 dataset.
As shown in \tabref{tab:Affinity}, in comparison to the random masking perturbation method utilized in DetectGPT, we observe a 1.92\% and 0.49\% increase in Affinity when employing the selective masking method.
Additionally, the mask-filling method yields affinity improvements of 3.38\% and 1.32\% for RoBERTa and \modelname{} models, respectively. 
These results illustrate that the Selective Multi-Strategy Perturbation method effectively preserves more distinguishable features between MGTs and HWTs.

\begin{table}[h]
\renewcommand{\arraystretch}{1.2}
\resizebox{\columnwidth}{!}{
\begin{tabular}{lccccc}
\toprule
\textbf{Model}& \textbf{RoBERTa} &&\textbf{\modelname{}}\\ \midrule
Random Mask $_{\textbf{DetectGPT}}$& -2.64 &&-1.64\\ 
Selective Mask $_{\textbf{\modelname{}}}$& -0.72&& -1.15\\ 
Mask-Filling $_{\textbf{DetectGPT}}$& -4.72&& -2.66\\ 
Mask-Filling $_{\textbf{\modelname{}}}$& -1.34&&-1.34 \\ \bottomrule
\end{tabular}
}
\caption{
Affinity of DetectGPT's and \modelname{}'s masking strategy on RoBERTa and \modelname{}.
}
\label{tab:Affinity}
\end{table}

\noindent\textbf{Analysis on Diversity}
Conversely, diversity assesses the range and variability of perturbed data, utilizing metrics Dist-1 and Dist-2 \cite{celikyilmaz2020evaluation}.
Here, we use three common perturbation methods to demonstrate the importance of not arbitrarily changing important tokens and the significance of select masks.
(1) Token Substitution (TS, \textcite{zhang2015character}), replaces tokens with synonyms from WordNet \cite{miller-1992-wordnet}; 
(2) SwitchOut (SO, \textcite{wang2018switchout}), uniformly samples and randomly substitutes from the vocabulary of test samples; 
and (3) Two-stage (TWs, \textcite{wei2021few}) trains the mask-filling model on the original data.

The ideal perturbation result is to have high Affinity scores while ensuring high Diversity scores \cite{celikyilmaz2020evaluation}. 
As shown in Table~\ref{tab:Diversity}, through Selective Strategy Perturbation, 
models achieve better diversity with high distribution shifts. 
And the overall improvement in Affinity by over 18\% also shows greater diversity than the original data.
The above results demonstrate the superiority of our perturbation method. 

\begin{table}[h]
\renewcommand{\arraystretch}{1.1}
\resizebox{\columnwidth}{!}{
\scriptsize
\begin{tabular}{lcccccc}
\toprule
\textbf{Method}&& \textbf{Affinity} &&\textbf{Dist-1}&&\textbf{Dist-2}\\ \midrule
TS&& -20.00 &&3.38&&43.43\\ 
TO&& -22.06&& 6.81&&53.61\\ 
TWs&& -21.13&& 3.24&&41.85\\ 
Original&& - &&8.70&&50.32\\ 
\midrule
\modelname{}&& -1.34&&15.59&&57.01 \\ \bottomrule
\end{tabular}
}
\caption{
Affinity and Diversity on GPT-2 datasets.
}
\label{tab:Diversity}
\end{table}

\subsubsection{Analysis on Selective Strategies}
In this section, we compare various strategies for selection in \modelname{}.
Beyond the \modelname{}'s importance-based perturbation method and random perturbation method (DetectGPT), we experiment with two other perturbation strategies: rank-based perturbation and keyword-based perturbation. In rank-based perturbation, we use the rescaled rank of next-token probability on GPT2-medium as the weight for perturbation position selection. In keyword-based perturbation, we prevent changes in the keywords extracted by the VLT-5 model \cite{Pęzik2022Keyword} during perturbation.
As shown in Table~\ref{tab:selectivefuther}, the experimental results of selective perturbation outperform the random perturbation method by 1.20\%, 2.04\%, and 2.49\% in average accuracy on the 64-example GPT2 dataset. And the importance-based strategy is the highest.
\begin{table}[h]
\renewcommand{\arraystretch}{1.2}
\resizebox{\columnwidth}{!}{
\begin{tabular}{lcccccc}
\toprule
\textbf{Method}&\textbf{Random} & \textbf{Prob. Rank} & \textbf{Keyword} &\textbf{Importance} \\ \midrule
Yake & 76.05$_{\text{1.83}}$ &77.35$_{\text{0.73}}$ &78.55$_{\text{1.65}}$ &
{\textbf{78.92}$_{\text{\textbf{1.14}}}$}\\ 
Perplexity & 75.53$_{\text{1.14}}$ &76.63$_{\text{1.03}}$ &77.11$_{\text{1.80}}$ &
{\textbf{77.63}$_{\text{\textbf{1.30}}}$}\\ 

\bottomrule
\end{tabular}
}
\caption{
Different strategies for perturbation and token-level weighting, namely Random (DetectGPT), Prob. Rank (GPT2-medium), Keyword (VLT-5), Importance (\modelname{}).
}
\label{tab:selectivefuther}
\end{table}

Further, we test the mask-filling failure ratio across the above strategies to interpret our outperformance. We find that the random strategy leads to more masking-filling failures than selective ones, which cause execution errors. 
Results in Table~\ref{tab:failure-ratio} indicate that selective strategy based on token importance performs the best, decreasing the failure ratio by 3.64\% than random. 
\begin{table}[h]
\renewcommand{\arraystretch}{1.2}
\resizebox{\columnwidth}{!}{
\begin{tabular}{lcccccc}
\toprule
\textbf{Method}& \textbf{Random} & \textbf{Prob. Rank} & \textbf{Keyword} &\textbf{Importance} \\ \midrule
Ratio (\%) & 9.20 &7.83&7.80 &5.56\\ 
\bottomrule
\end{tabular}
}
\caption{
Mask-filling failure ratio of different perturbation strategies.
}
\label{tab:failure-ratio}
\end{table}

\subsubsection{Generalization on Mask-Filling Models}

\begin{figure}[h]
  \centering
  \includegraphics[width=\columnwidth]{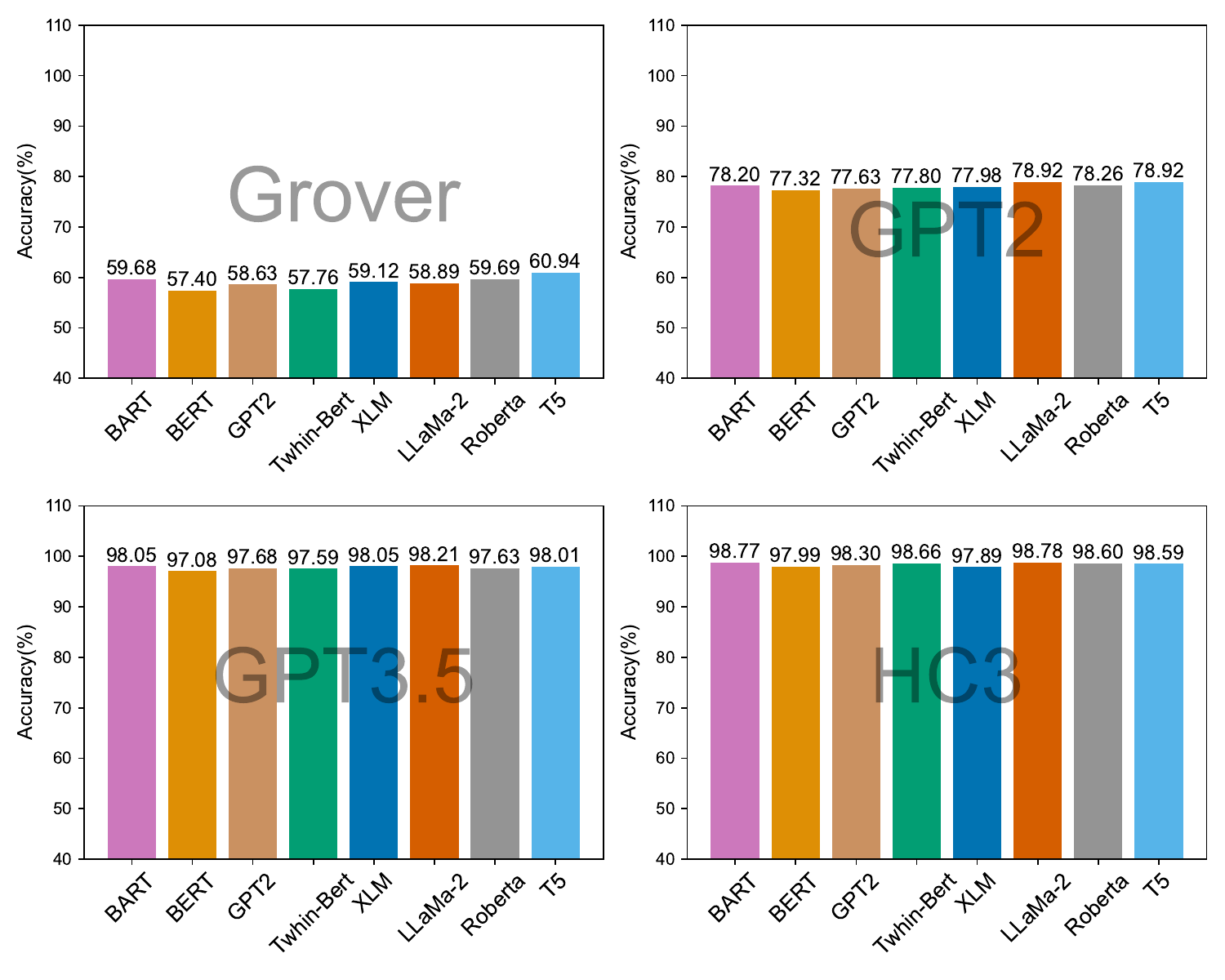}
  \caption{Result of generalizing on various mask-filling models.}
  \label{fig:mask-filling} 
\end{figure}

We study the influence of various mask-filling models on the performance of \modelname{}, including Bert (110M; \textcite{devlin2018bert}), Bart (139M; \textcite{lewis2019bart}), GPT-2 (380M; \textcite{radford2019language}), Twhin-bert (279M; \textcite{zhang2023twhin}), XLM (279M; \textcite{conneau2019unsupervised}), XLNet (110M; \textcite{yang2019xlnet}), RoBERTa (125M; \textcite{liu2019roberta}), and LLaMA-2 (7B; \textcite{touvron2023llama2}).
As depicted in \figref{fig:mask-filling}, the results of all mask-filling models surpass the baseline in terms of accuracy.
Furthermore, the fluctuation of \modelname{}'s performance across different mask-filling models is relatively slight.
It confirms that \modelname{} is not reliant on a specific filling model, showing great generalization capability.
The remaining full experimental results of different mask-filling models are in Appendix~\ref{app:maskfilling}.

\subsubsection{Generalization on Data}

\noindent\textbf{Cross-domain.}
We evaluate \modelname{} on the HC3 dataset crossing three QA domains, namely Medicine, Finance, and Computer Science. The meta-information details are in Appendix \ref{app:metadetal}.
For the three domains of data, we use one of them as training data (64-shot), and the remaining domains of data as testing data.
The results in \tabref{tab:cross-domin} show that \modelname{} is more effective than the best baseline and SOTA method on average. For example, compared to Roberta, \modelname{} outperforms by 4.61\% in three domains on average. 
And \sysname maintains a 1.63\% higher accuracy on average than SOTA DetectGPT. 

\begin{table}[h]
\renewcommand{\arraystretch}{1.2}
\resizebox{\columnwidth}{!}{
\begin{tabular}{lcccccc}
\toprule
\textbf{Domain}& \textbf{Medicine} & \textbf{Finance} &\textbf{Comp. Sci.}&\textbf{Average}\\ \midrule
RoBERTa & { 62.97$_{\text{4.09}}$}& { 86.08$_{\text{3.63}}$} &{ 90.64$_{\text{5.07}}$}&79.90\\ 
DetectGPT & \textbf{80.48}& 85.17 & 82.98 &82.88\\

\modelname{} & { 70.86$_{\text{7.83}}$}& {\textbf{89.34}$_{\text{\textbf{2.93}}}$}&{\textbf{93.32}$_{\text{\textbf{3.64}}}$}&\textbf{84.51}\\ 
\bottomrule
\end{tabular}
}
\caption{
Results of cross-domain in terms of accuracy.
}
\label{tab:cross-domin}
\end{table}

\noindent\textbf{Cross-generator.}
 We generalize \modelname{} between News articles (GPT3.5 dataset) and QA answers (HC3 dataset) on the 64-shot settings. 
As shown in \tabref{tab:cross-generator}, when the GPT-3.5 dataset is the training set, \modelname{} outperforms by 10.21\%; and when the HC3 dataset is the training set, \modelname{} outperforms by 6.98\% to the best competitor. 
\begin{table}[h]
\renewcommand{\arraystretch}{1.2}
\resizebox{\columnwidth}{!}{
\begin{tabular}{lcccccc}
\toprule
\textbf{Dataset}& \textbf{GPT3.5$\rightarrow$HC3} &  \textbf{HC3$\rightarrow$GPT3.5} &\textbf{Average}\\ \midrule
RoBERTa & { 64.60$_{\text{1.96}}$}& { 62.67$_{\text{2.41}}$}&63.64\\ 
DetectGPT & 77.11& 72.66&74.89\\ 
\modelname{} & {\textbf{78.79}$_{\text{\textbf{8.19}}}$}&{\textbf{72.87}$_{\text{\textbf{6.06}}}$}&\textbf{75.83}\\ 
\bottomrule
\end{tabular}
}
\caption{
Results of cross-generator in terms of accuracy.
}
\label{tab:cross-generator}
\end{table}

\subsubsection{Detecting Shorter Texts}

\begin{figure}[h]
  \centering
  \includegraphics[width=\columnwidth]{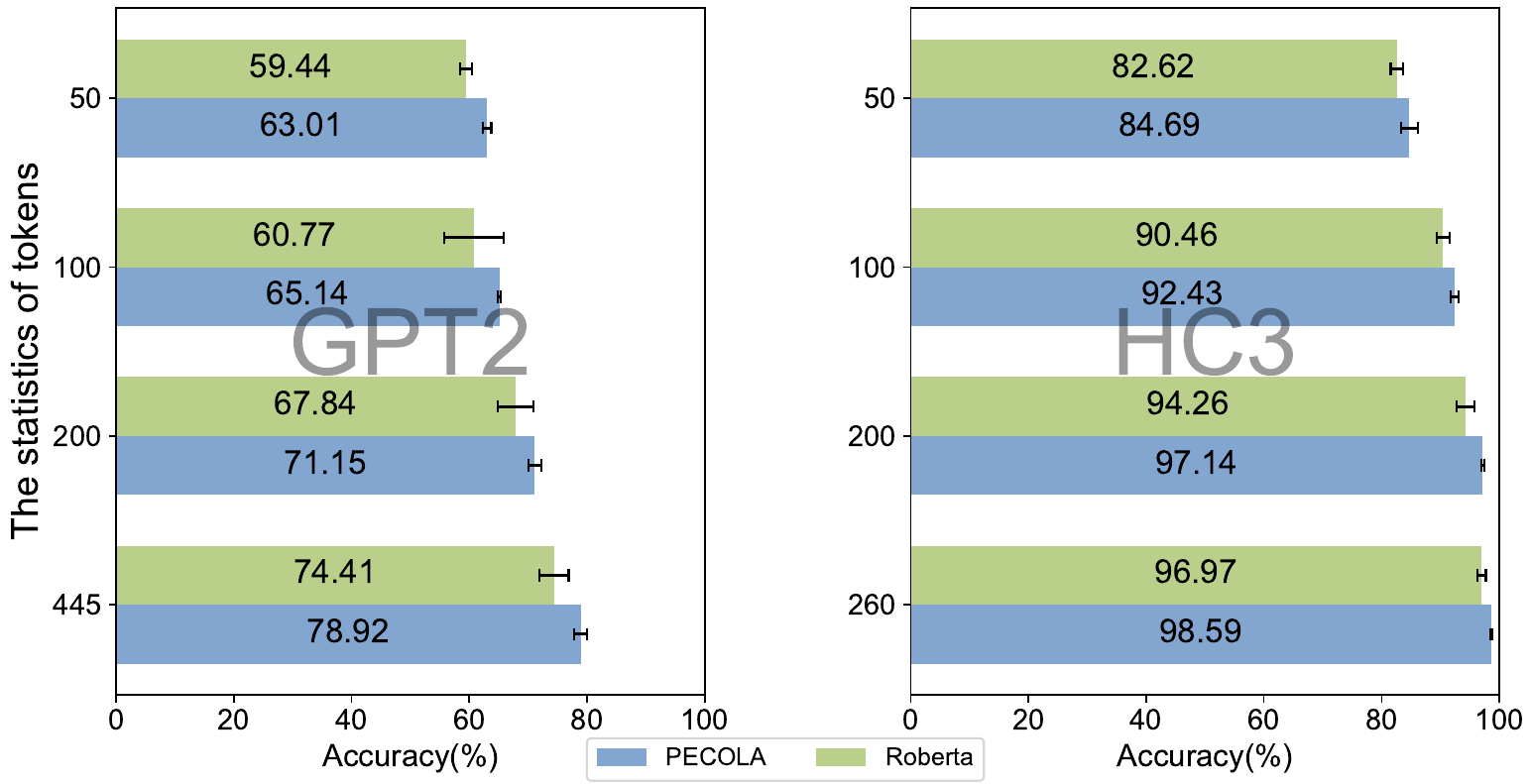}
  \caption{Performance of \modelname{} and RoBERTa to detect shorter texts. The average token number of the original GPT-2 and HC3 datasets are 445 and 260.}
  \label{fig:short} 
\end{figure}

To examine the efficiency of \modelname{} to detect the short MGTs, we chunk the samples of GPT-2 and HC3 datasets into segments of 50, 100, and 200 tokens.
As shown in \figref{fig:short}, \modelname{} consistently outperforms RoBERTa, with an average accuracy outperformance of 4.16\% and 2.13\% on the GPT-2 and HC3 datasets.
And the relative performance decrease of \modelname{} while the length shrinking is much less than RoBERTa.

\section{Conclusion}
In this paper, we introduce \modelname{}, a novel machine-generated text detection method that effectively bridges and integrates metric-based and fine-tuned detectors for MGT detection.
To relieve the information loss caused by the random masking used in DetectGPT, we present a token-level selective strategy perturbation method. 
To better distinguish meaningful recombination spaces and reduce reliance on the mask-filling models, we present a token-level weighted multi-pairwise contrastive learning method. 
In few-shot settings, experimental results show that \modelname{} significantly enhances the performance of PLMs in MGT detection.
Subsequent analytical experiments validate \modelname{}'s effectiveness, robustness, generalization, and capability in detecting short texts.

\section*{Acknowledgements}
We thank all the anonymous reviewers and the area chair for their helpful feedback, which aided us in greatly improving the paper.
This work is supported by National Key R\&D Program (2023YFB3107400), National Natural Science Foundation of China (62272371, 62103323, U21B2018, 62161160337,  U20B2049), Initiative Postdocs Supporting Program (BX20190275, BX20200270), China Postdoctoral Science Foundation (2019M663723, 2021M692565), Fundamental Research Funds for the Central Universities under grant (xzy012024144), and Shaanxi Province Key Industry Innovation Program (2021ZDLGY01-02).

\section*{Limitations}
In this work, we focus on MGT detection in few-shot learning settings. 
The next phase will involve a more comprehensive performance comparison based on full datasets.
Secondly, our method mentions the score threshold, if the threshold is too high or too low, it will not serve the purpose of perturbation. 
How to automate and flexibly design a strict threshold is also a direction for our next phase of improvement.
Thirdly, for short texts, our perturbation method faces similar limitations, as it is difficult to extract the most relevant keywords. 
Thus, perturbation introduces more uncontrollable noise, which poses a challenge for us to address in the future.
Fourth, We hope that the present work can inspire future applications in fields like machine-generated images and videos, creating a universal approach to apply in the direction of machine generation.

\section*{Ethics Statement}

\modelname{} aims to help users use our method to more reasonably and accurately identify MGT. Our goal is to develop a universal method applicable to other fields such as images and audio, and inspire the advancement of the stronger detector of MGTs and prevent all potential negative uses of language models. 
We do not wish our work to be maliciously used to counter detectors. The datasets mentioned in this paper are all public.

\bibliography{custom}
\bibliographystyle{acl_natbib}

\appendix

\clearpage

\section{Implementation Details}

This part mentions the hyperparameter settings and meta-information of the HC3 dataset. 

\subsection{Hyperparameter Details}
\label{app:details}
Experiments evaluating competitors and \modelname{} follow the setting of CoCo \cite{liu-etal-2023-coco}. The hyperparameter settings of all the methods in the experiment as shown in Table~\ref{tab:Detail}. We randomly select 10 different seeds for experiments, and report average test accuracy and F1-score.

\begin{table}[ht]
\centering
\renewcommand\arraystretch{1.3}
\tabcolsep=3pt
{\fontsize{11pt}{12pt}\selectfont
\begin{tabular}{lc} 
\toprule
\textbf{Parameter} & \textbf{Value} \\
\midrule
Training Epochs & 30 \\
Optimizer & AdamW \\ 
Learning rate & 1e-5 \\
Weight Decay & 0.01 \\
Batch Size & 16 \\ 
Mask Gap &2 \\ 
Mask Proportion  &10\%\\ 
Score threshold & 0.4 \\ 
Pre-trained model  \ & \ RoBERTa-base \\                                 
\bottomrule
\end{tabular}
}
\caption{Implementation details of hyperparameters.}
\label{tab:Detail}
\end{table}

\subsection{Dataset Meta Information}
We evaluate \modelname{} effectiveness from domains and generators on the HC3 dataset, which primarily includes Medicine, Finance, and Computer Science domain QA, as shown in Table~\ref{tab:Meta}.
\label{app:metadetal}
\begin{table}[ht]
\renewcommand{\arraystretch}{1.2}
\begin{tabular}{lccccc}
\toprule
\textbf{Domain}& Medicine & Finance & Comp. Sci.\\ \midrule
Size &2585&8436 &1684\\ 

\bottomrule
\end{tabular}
\caption{
Meta-information of the HC3 dataset.
}
\label{tab:Meta}
\end{table}

\section{Effect of Hyperparameters}

In \modelname{}, the primary hyperparameters include the mask proportion, mask gap of perturbation, and score threshold. 
The perturbation proportion refers to the mask rate in the texts.
The perturbation mask gap ensures that several tokens following a masked token remain unmasked, and score threshold to control the number of Most Relevant Keywords.

\subsection{Perturbation Proportion and Mask Gap}
\begin{figure}[ht]
  \centering
  \includegraphics[width=\columnwidth]{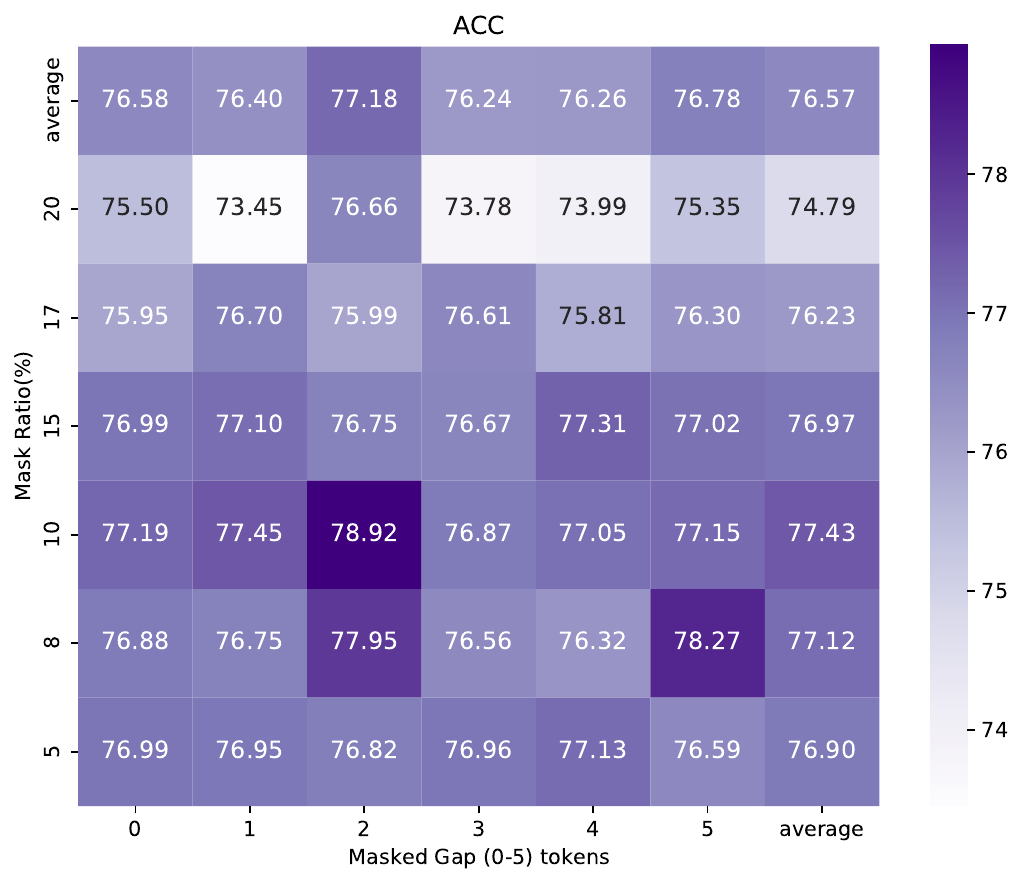}
  \caption{Impact of varying the number of perturbations and mask gap in \modelname{}, we use T5-large \cite{raffel2020exploring} as the mask-filling model. For each combination, we conduct tests on ten randomly select seeds.}
  \label{fig:span1} 
\end{figure}

We evaluate the impact of different perturbation ratios and mask gap on accuracy, and perform a minor scan in a few-shot learning settings with a set of mask proportions \{5, 8, 10, 15, 17, 20\} and mask gap \{0, 1, 2, 3, 4, 5\}, average the results for each combination of parameters. 
And a mask gap of 2 and a perturbation ratio of 10\% achieve the maximum average values.
As shown in \figref{fig:span1}, it is found that the combination of a mask gap of 2 and a mask proportion of 10\% yielded the best results, on the 64-example GPT-2 dataset.

\subsection{Score Threshold}
\begin{figure*}
  \centering
  \includegraphics[width=0.8\textwidth]{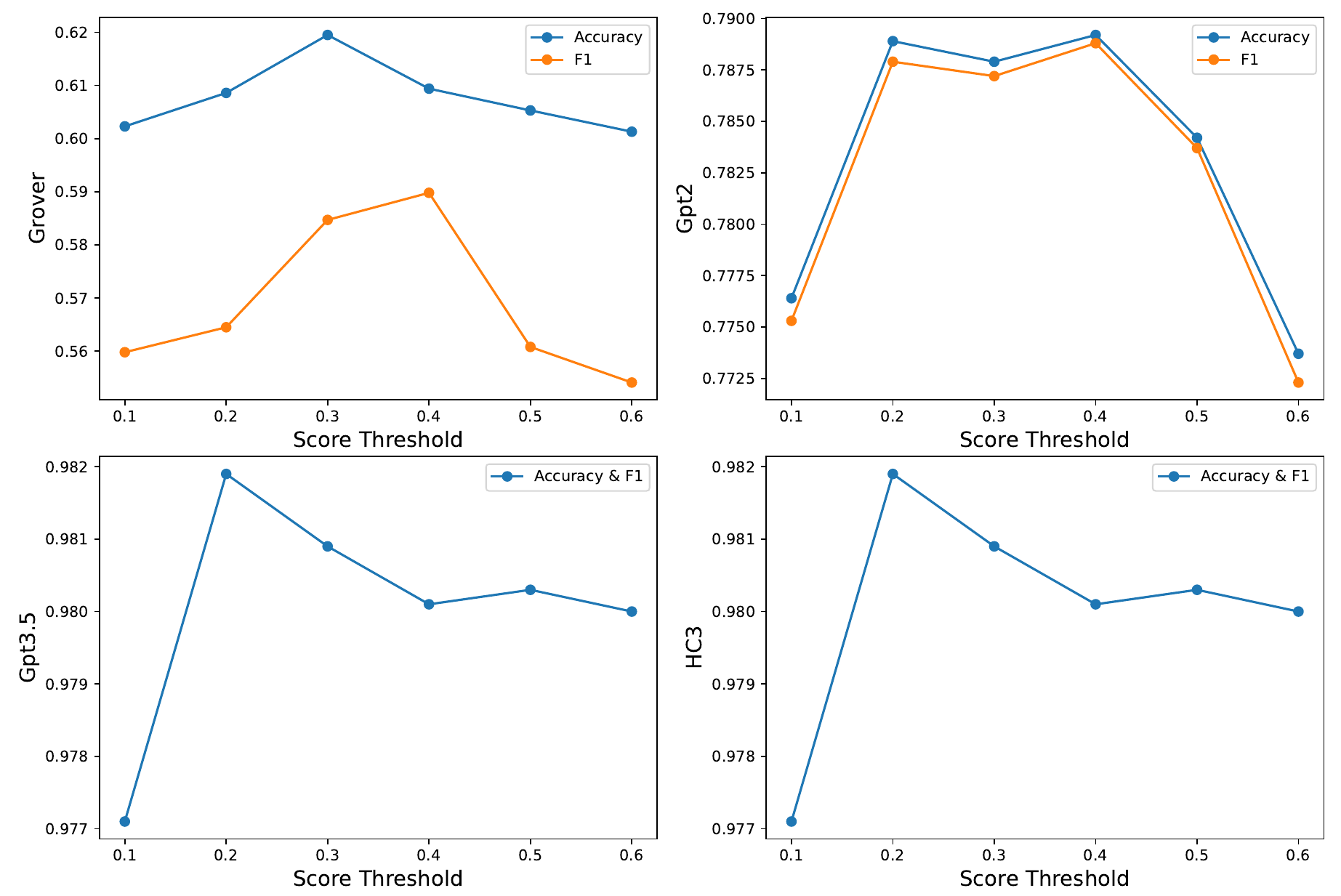}
  \caption{Effect of score threshold on model performance. In the GPT3.5 and HC3 datasets (sub-figure 3 and 4), accuracy and F1-score coincide.}
  \label{fig:threold} 
\end{figure*}

In the main experiment, all datasets use a common score threshold of 0.4, and it may not be the best choice for different datasets, because with the change in data type and text length, the gold keywords often vary. 
Therefore, as shown in \figref{fig:threold}, we discuss the performance changes of four datasets with different score threshold in few-shot learning settings. An excessively high score threshold results in too many most relevant keywords, failing to effectively perturb the data, hence not significantly improving accuracy.
 Similarly, a too low score threshold can lead to more random perturbations. Therefore, the selection of the score threshold should be stringent.
 

\section{Efficiency of \sysname}

\subsection{Scalability of Base Models at Different Scales}
We adopt Pythia \cite{Biderman2023Pythia:} as the base model of \modelname{} with different scales, \ie 70M, 160M, 410M, 1B, and 1.4B. We train and do experiments on one NVIDIA A100 GPU, and the performance and time consumption are in Table~\ref{tab:scalability}. With the increase in model size, both accuracy and F1-score show upward trends, while the time increase is linear, which is reasonable.
\label{app:scalability}
\begin{table*}[ht]
\centering
\renewcommand{\arraystretch}{1.1}
\begin{tabular}{lcccccc}
\toprule
\textbf{Model}& \textbf{70M} & \textbf{160M} &\textbf{410M}&\textbf{1B}&\textbf{1.4B}\\ \midrule
Acc & { 58.42$_{\text{0.70}}$}& { 63.66$_{\text{0.17}}$} &{ 71.07$_{\text{1.63}}$}&{72.13$_{\text{1.63}}$}&{74.05$_{\text{1.77}}$}\\ 
F1 & { 58.03$_{\text{0.79}}$}& { 63.54$_{\text{0.28}}$} &{ 70.87 $_{\text{1.92}}$}&{71.75$_{\text{2.67}}$}&{73.85$_{\text{1.55}}$}\\ 

Per epoch & 16s& 34s & 85s &97s& 113s\\
Single data & 2.2ms & 7.0ms & 13.8ms &14.1ms&16.6ms\\

\bottomrule
\end{tabular}
\caption{
Results of fine-tuning \modelname{} with Pythia models of various scales, on the 64-example GPT2 dataset.
We also demonstrate the training time per epoch and the single data test time.
}
\label{tab:scalability}
\end{table*}

\subsection{Impact of the Chosen Mask-filling Models}

This section shows the full experimental results of different mask-filling models, as shown in Table~\ref{tab:full}, the experimental results confirm the same outcomes as in the few-shot learning settings, where the T5 filling model does not perform the best across all datasets.
All the above models are obtained from huggingface transformers \cite{wolf2020transformers}. And we do not intervene in the temperature sampling of the mask-filling model, setting it all to 1.
\label{app:maskfilling}
\begin{table*}[t]
    \centering
    \renewcommand\arraystretch{1.2}
    \resizebox{\textwidth}{!}{
    \begin{tabular}{cccccccccccp{1.25cm}<{\centering}}\hline
{\textbf{Dataset}}
& {\textbf{Method}}
& {\textbf{Shot}}
& {\textbf{\textit{BART}}}
& {\textbf{\textit{Bert}}}
& {\textbf{\textit{GPT-2}}}
& {\textbf{\textit{Twhin Bert}}} 
& {\textbf{\textit{XLM}}}
& {\textbf{\textit{XLNet}}}
& {\textbf{\textit{RoBERTa}}}  
& {\textbf{\textit{T5}}} \\
    \toprule
              
            \multirow{6}{*}{\textbf{\rotatebox{90}{Grover}}} &
            \multirow{2}{*}{\textit{Acc}}&128 &62.04$_{\text{2.51}}$ &61.55$_{\text{1.74}}$ &62.82$_{\text{1.24}}$ &61.00$_{\text{2.20}}$ &61.82$_{\text{0.82}}$ &60.16$_{\text{0.43}}$ &63.10$_{\text{1.76}}$ &\textbf{63.60}$_{\text{\textbf{1.71}}}$ \\
            & &512 &72.24$_{\text{1.54}}$ &71.67$_{\text{1.04}}$ &72.62$_{\text{1.12}}$ &72.78$_{\text{1.14}}$ &72.13$_{\text{0.64}}$ &72.72$_{\text{1.03}}$ &\textbf{73.25}$_{\text{\textbf{0.84}}}$ &73.12$_{\text{0.84}}$ \\ 
            \cline{3-11} 

            &\multirow{2}{*}{\textit{F1}}&128 &57.80$_{\text{1.28}}$ &57.60$_{\text{1.93}}$ &58.55$_{\text{0.80}}$ &56.74$_{\text{0.48}}$ &57.60$_{\text{0.92}}$ &56.62$_{\text{0.64}}$ &58.29$_{\text{1.12}}$ &\textbf{58.98}$_{\text{\textbf{1.58}}}$
            \\
            & &512 &66.25$_{\text{2.34}}$ &65.56$_{\text{1.76}}$ &66.72$_{\text{2.00}}$ &68.49$_{\text{1.04}}$ &66.38$_{\text{2.21}}$ &67.50$_{\text{2.61}}$ &67.49$_{\text{1.68}}$ &\textbf{\textbf{68.24}$_{\text{\textbf{1.64}}}$}
            \\
            \cline{3-11} 
            
            &\multirow{2}{*}{\textit{Recall}}&128 &58.03$_{\text{0.99}}$ &57.91$_{\text{2.08}}$ &\textbf{58.72}$_{\text{\textbf{0.87}}}$ &57.18$_{\text{0.86}}$ &57.78$_{\text{1.04}}$ &57.00$_{\text{0.80}}$ &58.31$_{\text{0.99}}$ &57.89$_{\text{1.44}}$
            \\
            & &512 &65.85$_{\text{2.66}}$ &64.87$_{\text{1.71}}$ &66.01$_{\text{2.06}}$ &\textbf{68.11}$_{\text{\textbf{1.16}}}$ &65.87$_{\text{2.46}}$ &67.05$_{\text{3.04}}$ &66.73$_{\text{1.68}}$ &66.51$_{\text{1.64}}$
            \\ 
            \hline

            \multirow{6}{*}{\textbf{\rotatebox{90}{GPT-2}}} &
            \multirow{2}{*}{\textit{Acc}}&128 &82.16$_{\text{1.04}}$ &80.77$_{\text{0.48}}$ &82.42$_{\text{1.05}}$ &82.17$_{\text{0.40}}$ &81.15$_{\text{0.31}}$ &81.26$_{\text{0.36}}$ &81.27$_{\text{1.20}}$ &\textbf{82.58}$_{\text{\textbf{0.49}}}$ \\
            & &512 &85.41$_{\text{0.66}}$ &85.43$_{\text{0.53}}$ &85.52$_{\text{0.57}}$  &85.72$_{\text{0.39}}$ &85.10$_{\text{0.27}}$ &85.13$_{\text{0.60}}$ &85.75$_{\text{0.55}}$ &\textbf{85.75}$_{\text{\textbf{0.69}}}$ \\ 
            \cline{3-11} 

            &\multirow{2}{*}{\textit{F1}}&128 &82.12$_{\text{1.07}}$ &80.67$_{\text{0.54}}$ &82.38$_{\text{1.08}}$ &82.12$_{\text{0.38}}$ &81.11$_{\text{0.34}}$ &81.24$_{\text{0.37}}$ &81.16$_{\text{1.27}}$ &\textbf{82.54}$_{\text{\textbf{0.51}}}$
            \\
            & &512 &85.40$_{\text{0.67}}$ &85.41$_{\text{0.53}}$ &85.72$_{\text{0.70}}$ &85.72$_{\text{0.39}}$ &85.10$_{\text{0.27}}$ &85.13$_{\text{0.60}}$ &\textbf{85.75}$_{\text{\textbf{0.55}}}$ &85.72$_{\text{0.70}}$
            \\
            \cline{3-11} 
            
            &\multirow{2}{*}{\textit{Recall}}&128 &82.15$_{\text{1.05}}$ &80.75$_{\text{0.48}}$ &82.01$_{\text{0.68}}$ &82.17$_{\text{0.40}}$ &81.15$_{\text{0.31}}$ &81.26$_{\text{0.36}}$ &81.25$_{\text{1.20}}$ &\textbf{82.57}$_{\text{\textbf{0.49}}}$
            \\
            & &512 &85.41$_{\text{0.66}}$ &85.43$_{\text{0.53}}$ &\textbf{85.80}$_{\text{\textbf{0.27}}}$ &85.72$_{\text{0.39}}$ &85.10$_{\text{0.27}}$ &85.13$_{\text{0.60}}$ &85.75$_{\text{0.55}}$ &85.52$_{\text{0.57}}$ 
            \\ 
            \hline

            \multirow{6}{*}{\textbf{\rotatebox{90}{GPT-3.5}}} &
            \multirow{2}{*}{\textit{Acc}}&128 &98.24$_{\text{0.16}}$ &98.09$_{\text{0.25}}$ &98.09$_{\text{0.10}}$ &98.11$_{\text{0.11}}$ &97.98$_{\text{0.14}}$ &98.13$_{\text{0.08}}$ &98.01$_{\text{0.18}}$ &\textbf{98.63}$_{\text{\textbf{0.32}}}$\\
            & &512 &99.19$_{\text{0.13}}$ &99.05$_{\text{0.15}}$ &99.13$_{\text{0.17}}$ &98.89$_{\text{0.21}}$ &98.88$_{\text{0.21}}$ &\textbf{99.23}$_{\text{\textbf{0.26}}}$ &99.16$_{\text{0.14}}$ &99.15$_{\text{0.11}}$ \\
            \cline{3-11} 

            &\multirow{2}{*}{\textit{F1}}&128 &98.24$_{\text{0.16}}$ &98.09$_{\text{0.25}}$ &98.09$_{\text{0.10}}$ &98.11$_{\text{0.11}}$ &97.98$_{\text{0.14}}$ &98.13$_{\text{0.08}}$ &98.01$_{\text{0.18}}$ &\textbf{98.63}$_{\text{\textbf{0.32}}}$\\
            & &512 &99.19$_{\text{0.13}}$ &99.05$_{\text{0.15}}$ &99.13$_{\text{0.17}}$ &98.89$_{\text{0.21}}$ &98.88$_{\text{0.21}}$ &\textbf{99.23}$_{\text{\textbf{0.26}}}$ &99.16$_{\text{0.14}}$ &99.15$_{\text{0.11}}$ \\
            \cline{3-11} 
            
            &\multirow{2}{*}{\textit{Recall}}&128 &98.24$_{\text{0.16}}$ &98.09$_{\text{0.25}}$ &98.09$_{\text{0.10}}$ &98.11$_{\text{0.11}}$ &97.98$_{\text{0.14}}$ &98.13$_{\text{0.08}}$ &98.01$_{\text{0.18}}$ &\textbf{98.63}$_{\text{\textbf{0.32}}}$\\
            & &512 &99.19$_{\text{0.13}}$ &99.05$_{\text{0.15}}$ &99.13$_{\text{0.17}}$ &98.89$_{\text{0.21}}$ &98.88$_{\text{0.21}}$ &\textbf{99.23}$_{\text{\textbf{0.26}}}$ &99.16$_{\text{0.14}}$ &99.15$_{\text{0.11}}$ \\
            \hline
            
            \multirow{6}{*}{\textbf{\rotatebox{90}{HC3}}} &
            \multirow{2}{*}{\textit{Acc}}&128 &98.63$_{\text{0.18}}$ &98.03$_{\text{0.40}}$ &98.59$_{\text{0.16}}$ &98.58$_{\text{0.22}}$ &98.24$_{\text{0.09}}$ &98.35$_{\text{0.12}}$ &\textbf{98.79}$_{\text{\textbf{0.32}}}$ &98.06$_{\text{0.12}}$ \\
            & &512 &98.82$_{\text{0.35}}$ &98.45$_{\text{0.21}}$ &98.96$_{\text{0.25}}$ &98.83$_{\text{0.24}}$ &98.80$_{\text{0.38}}$ &98.80$_{\text{0.30}}$ &99.02$_{\text{0.23}}$ &\textbf{99.14}$_{\text{\textbf{0.15}}}$ \\
            \cline{3-11} 

            &\multirow{2}{*}{\textit{F1}}&128 &98.63$_{\text{0.18}}$ &98.03$_{\text{0.40}}$ &98.59$_{\text{0.16}}$ &98.58$_{\text{0.22}}$ &98.24$_{\text{0.09}}$ &98.35$_{\text{0.12}}$ &\textbf{98.79}$_{\text{\textbf{0.32}}}$ &98.06$_{\text{0.12}}$ \\
            & &512 &98.82$_{\text{0.35}}$ &98.45$_{\text{0.21}}$ &98.96$_{\text{0.25}}$ &98.83$_{\text{0.24}}$ &98.80$_{\text{0.38}}$ &98.80$_{\text{0.30}}$ &99.02$_{\text{0.23}}$ &\textbf{99.14}$_{\text{\textbf{0.15}}}$ \\
            \cline{3-11} 
            
            &\multirow{2}{*}{\textit{Reacall}}&128 &98.63$_{\text{0.18}}$ &98.03$_{\text{0.40}}$ &98.59$_{\text{0.16}}$ &98.58$_{\text{0.22}}$ &98.24$_{\text{0.09}}$ &98.35$_{\text{0.12}}$ &\textbf{98.79}$_{\text{\textbf{0.32}}}$ &98.63$_{\text{0.32}}$ \\
            & &512 &98.82$_{\text{0.35}}$ &98.45$_{\text{0.21}}$ &98.96$_{\text{0.25}}$ &98.83$_{\text{0.24}}$ &98.80$_{\text{0.38}}$ &98.80$_{\text{0.30}}$ &99.02$_{\text{0.23}}$ &\textbf{99.15}$_{\text{\textbf{0.11}}}$ \\
            \bottomrule
              
\end{tabular}
}
    \caption{\label{citation-guide}The full MGT detection performance of different mask-filling models on four datasets. We use the model version with the same level model size, \,  i.e. {} base version for most models.
    }
\label{tab:full}
\end{table*}

\begin{table*}[t!]
\centering
\renewcommand\arraystretch{1.2}
\resizebox{\textwidth}{!}{
\begin{tabular}{cccccccccccp{1.25cm}<{\centering}}\hline
{\textbf{Dataset}}
& {\textbf{Shot}}
& {\textbf{Metric}}
& {\textbf{\textit{RoBERTa}}}
& {\textbf{\textit{GLTR}}}
& {\textbf{\textit{CE+SCL}}}
& {\textbf{\textit{CE+Margin}}} 
& {\textbf{\textit{IT:Clust}}}
& {\textbf{\textit{CoCo}}}
& {\textbf{\textit{DetectGPT}}}  
& {\textbf{\textit{Fast-Detect.}}}
& {\textbf{\textit{\modelname{}}}}  \\
    \hline
              
            \multirow{2}{*}{\textbf{Grover}} &
            \multirow{2}{*}{\textbf{10000}}&
            \multirow{1}{*}{\textit{Acc}}
            &86.13$_{\text{0.47}}$ &60.40 &\textbf{86.57}$_{\text{\textbf{0.44}}}$
            &86.25$_{\text{0.81}}$ &72.65$_{\text{3.44}}$&85.23$_{\text{0.20}}$ &61.42&65.49&\textbf{86.70}$_{\text{\textbf{0.37}}}$ \\
            & &\multirow{1}{*}{\textit{F1}}
            &84.07$_{\text{0.91}}$ &59.82 &84.95$_{\text{0.56}}$ &\textbf{85.10}$_{\text{\textbf{1.27}}}$ &63.21$_{\text{5.02}}$&83.67$_{\text{0.56}}$ &54.28&63.29&\textbf{86.66}$_{\text{\textbf{0.33}}}$ \\

            \cline{3-12} 
            \hline

            \multirow{2}{*}{\textbf{GPT-2}} &
            \multirow{2}{*}{\textbf{10000}}&
            \multirow{1}{*}{\textit{Acc}}
            &89.56$_{\text{1.18}}$ &77.55 &90.19$_{\text{0.60}}$ &\textbf{90.30}$_{\text{\textbf{0.41}}}$ &81.65$_{\text{2.14}}$&89.78$_{\text{0.04}}$ 
            &78.74&80.06&\textbf{91.10}$_{\text{\textbf{0.09}}}$ \\
            & &\multirow{1}{*}{\textit{F1}}
            &89.51$_{\text{1.15}}$ &76.39 &90.15$_{\text{0.61}}$ &            \textbf{90.27}$_{\text{\textbf{0.40}}}$ &81.54$_{\text{3.20}}$&89.01$_{\text{0.07}}$ 
            &71.13&80.64&\textbf{91.10}$_{\text{\textbf{0.10}}}$ \\

            \cline{3-12} 
            \hline

            \multirow{2}{*}{\textbf{GPT-3.5}} &
            \multirow{2}{*}{\textbf{7000}}&
            \multirow{1}{*}{\textit{Acc}}
            &99.89$_{\text{0.03}}$ &93.50 &99.74$_{\text{0.04}}$ &\textbf{99.90}$_{\text{\textbf{0.03}}}$         
            &99.09$_{\text{0.31}}$&99.44$_{\text{0.12}}$ &90.80&94.72&\textbf{99.95}$_{\text{\textbf{0.01}}}$ \\
            & &\multirow{1}{*}{\textit{F1}}
            &99.89$_{\text{0.03}}$ &93.58 &99.74$_{\text{0.04}}$ 
            &\textbf{99.90}$_{\text{\textbf{0.03}}}$
            &99.09$_{\text{0.31}}$&99.44$_{\text{0.12}}$ &89.14&94.76&\textbf{99.95}$_{\text{\textbf{0.01}}}$ \\

            \cline{3-12} 
            \hline

            \multirow{2}{*}{\textbf{HC3}} &
            \multirow{2}{*}{\textbf{10000}}&
            \multirow{1}{*}{\textit{Acc}}
            &99.84$_{\text{0.08}}$ &98.39 &\textbf{99.89}$_{\text{\textbf{0.01}}}$ &99.86$_{\text{0.03}}$ &98.80$_{\text{0.67}}$&99.46$_{\text{0.24}}$ &95.13&98.32&\textbf{99.92}$_{\text{\textbf{0.01}}}$ \\
            & &\multirow{1}{*}{\textit{F1}}
            &99.84$_{\text{0.08}}$ &98.49 &\textbf{99.89}$_{\text{\textbf{0.01}}}$ &99.86$_{\text{0.03}}$ &98.80$_{\text{0.67}}$&99.46$_{\text{0.24}}$ &95.05&98.02&\textbf{99.92}$_{\text{\textbf{0.01}}}$ \\

            \cline{3-12} 
            \hline

\end{tabular}
}
    \caption{ Performance comparison of \modelname{} to baseline methods on the full datasets. The results are average values of 5 runs with different random seeds. Bold shows the best and second-best results within each column.}
\label{tab:full data}
\end{table*}

\subsection{Further Experiments on Full Datasets}
To demonstrate Pecola's superiority over the whole training set, we conduct a more in-depth test, as shown in Table~\ref{tab:full data}. We train the detector using 10,000 samples from the Grover, GPT-2, and HC3 datasets, and 7,000 samples from GPT-3.5 as our training sets. Comparatively, \modelname{} outperforms the second-best results in accuracy and F1-score by 0.13\% and 1.56\%, 0.80\% and 0.83\%, 0.05\% and 0.05\%, 0.03\% and 0.03\% respectively, across four datasets.
\label{app:fulldata}

\end{document}